\documentclass [10pt] {article}
\usepackage{amsmath}
\usepackage{amsfonts}
\usepackage{amssymb}
\usepackage{mathrsfs}
\usepackage{algorithm}
\usepackage{algorithmic}
\usepackage{placeins}

\usepackage [square,sort,comma,numbers] {natbib}

\usepackage{fancyhdr}
\usepackage{blindtext}

\usepackage{MnSymbol}

\usepackage[cal=boondox,scr=boondoxo]{mathalfa}

\usepackage{float}
\usepackage{subfig}
\usepackage{fancybox,graphicx}
\usepackage{subfig}
\usepackage{caption}
\usepackage{color}
\usepackage{authblk}

\usepackage[colorlinks]{hyperref}

\usepackage{accents}
\usepackage[titletoc,title]{appendix}
\usepackage{cite}

\usepackage{mathtools}

\usepackage[top=2in, bottom=1.5in, left=1in, right=1in]{geometry}

\usepackage{stackengine}

\title{From Single Agent to Multi-Agent: Improving Traffic Signal Control} 
 
\author[1]{Tislenko Maksim}
\author[2]{Kisilev Dmitrii}

\affil[1]{ITMO University, Saint-Petersburg, e-mail: makstislenko@gmail.com}
\affil[2]{HSE University, Moscow, e-mail: dkiseljov@hse.ru}

\begin{document}

    \maketitle

\begin{abstract}
Due to accelerating urbanization, the importance of solving the signal control problem increases. This paper analyzes various existing methods and suggests options for increasing the number of agents to reduce the average travel time. Experiments were carried out with 2 datasets. The results show that in some cases, the implementation of multiple agents can improve existing methods. For a fine-tuned large language model approach there's small enhancement on all metrics.
\end{abstract}

\section{Introduction}

The issue of traffic signal control is more relevant than ever. According to the UN report, "The urban population of the world has grown rapidly since 1950, having increased from 751 million to 4.2 billion in 2018. Asia, despite being less urbanized than most other regions today, is home to 54 \% of the world's urban population, followed by Europe and Africa (13\% each).\citep{un2018urbanization}". In 2014, traffic congestion cost Americans over \$ 160 billion in lost productivity and wasted over 3.1 billion gallons of fuel \citep{economist2014traffic}. Traffic congestion was also attributed to over 56 billion pounds of harmful CO2 emissions in 2011 \citep{schrank2015urban}. In the European Union, the cost of traffic congestion was equivalent to 1 \% of the entire GDP \citep{schrank2012urban}. Consequently, efficient TSC offers significant economic and environmental benefits, as well as improvements in societal well-being.

There're lots of different methods how to get less number of traffic congestions, allow people to reach their destination faster on average from reinforcement learning methods to algorithms based on large language models. In lots of these effective algorithms, multi-agent approach was not applied despite the fact that it showed good results as was proved in \citep{li2024agents}. 

In this article, we've created a multi-agent system where agents can utilize different approaches, including those based on Reinforcement Learning (RL) and Large Language Models (LLMs). Our focus is to adapt the method from a \citep{lai2024llmlight} to the domain of traffic signal control and then conduct experiments with multiple agents to observe and analyze their performance. Specifically, we hypothesize that the combination of RL or LLM-based approaches can improve traffic signal control by leveraging the strengths of both methodologies. By using a majority voting mechanism, we aim to aggregate the decisions of diverse agents to achieve a more robust and effective traffic signal control system. That's why we are attempting to implement the method of majority voting among several agents. The details of these methods and their implementation will be explained further in the article. 

The experiments will involve setting up a simulated traffic environment where different agents control traffic signals at various intersections.  We've conducted experiments on 2 datasets with 1, 5, or 10 agents and compared results for different numbers of agents. 

\section{Related work}

Existing research on Traffic Signal Control (TSC) can be categorized into three main approaches: traditional transportation methods \citep{koonce2008traffic, varaiya2013max, hunt1982scoot}, Reinforcement Learning (RL)-based techniques \citep{wei2019colight, wei2019presslight, oroojlooy2020attendlight, chen2020toward}, and methods based on transformers \citep{metagraph, lai2024llmlight, wu2023transformerlight}

Transportation methods focus on developing efficient heuristic algorithms that adapt traffic signal configurations dynamically based on real-time lane-level traffic conditions. However, these methods require significant manual design and human effort.

The advent of deep neural networks (DNNs) has led to the adoption of deep RL-based techniques, which have shown impressive performance in various traffic scenarios. Despite their success, RL-based methods face several challenges. They often struggle with generalizing to larger road networks or rare high-traffic situations, as their training data covers only a limited range of traffic scenarios. Additionally, RL-based methods lack interpretability due to the black-box nature of DNNs, making it difficult to understand the reasoning behind their control actions in specific traffic conditions.

Our research uses the MPLight algorithm that was introduced in \citep{chen2020toward}. It leverages the concept of pressure to coordinate multiple intersections. Pressure is defined as the difference between the queue lengths of incoming lanes at an intersection and the queue length of the downstream intersection's receiving lane. Chen et al. employed pressure as both the state and reward for a DQN agent, which is shared across all intersections, built on top of the FRAP \citep{zheng2019learning} model. The authors reported that this approach resulted in up to a 19.2\% improvement in travel times compared to the next best method, PressLight \citep{wei2019presslight}.

There are also a few methods based on transformers. For instance, in \citep{inproceedings} transformers are used to characterize traffic conditions better. With the shared Transformer-based cooperation mechanism, intersection controllers could adequately exploit spatial-temporal correlations and adaptively capture global traffic dynamics, guiding them to explore collaborative traffic strategies more efficiently. 

Also in \citep{lai2024llmlight} proposed a new method based on LLMs. In this study, LLMLight was presented, a novel framework that utilizes large language models (LLMs) as traffic signal control agents. Specifically, the framework begins by instructing the LLM with a knowledgeable prompt detailing real-time traffic conditions. By guiding LLMs to perform a human-like, step-by-step analysis of current traffic conditions, the intelligent agent can make optimal control decisions, thereby improving overall traffic efficiency at intersections. Extensive experiments on nine traffic flow datasets demonstrate that the framework equipped with LightGPT, a backbone training procedure for optimizing a traffic signal control-oriented LLM, significantly outperforms traditional methods, showcasing its exceptional effectiveness and generalization ability across various traffic scenarios. Leveraging the advanced generalization capabilities of LLMs, LLMLight engages a reasoning and decision-making process akin to human intuition for effective traffic control. This work showed that LLMs show results better than traditional RL and transportation methods in some cases. Also, it is worth noticing that LLM-powered traffic signal control not only shows superior performance across diverse traffic scenarios but also provides detailed explanations for each decision, contributing to a more comprehensible and accountable traffic control system. Thus, there is a reason to use LLM for traffic control problems and try to improve the performance of LLM using various methods.

There're lots of different ways to improve the performance of existing large language models. For instance \citep{chen2024selfplay} focus on improvement based on a fine-tuning method called Self-Play fIne-tuNing (SPIN), which starts from a supervised fine-tuned model. At the heart of SPIN lies a self-play mechanism, where the LLM refines its capability by playing against instances of itself. More specifically, the LLM generates its own training data from its previous iterations, refining its policy by discerning these self-generated responses from those obtained from human-annotated data. The method progressively elevates the LLM from a nascent model to a formidable one, unlocking the full potential of human-annotated demonstration data for SFT. Some of the recent studies focus on ensemble methods \citep{Wang2022SelfConsistencyIC, wan2024knowledge} and multiple LLM-Agents collaboration frameworks \citep{du, wu}. In these works, multiple LLM agents are used to improve the performance of LLMs. For instance, LLM-Debate \citep{du} employs multiple LLM agents in a debate form. The reasoning performance is improved by creating a framework that allows more than one agent to "debate" the final answer to arithmetic tasks. They show performance improvements compared to using one single agent. All in all, these researches use under the hood multiple LLMs to increase the performance. 

Our work is based on \citep{li2024agents}. In this paper simply via a sampling-and-voting method, the performance of large language models (LLMs) scales with the number of agents instantiated. Also, this method is orthogonal to existing complicated methods to further enhance LLMs, while the degree of enhancement is correlated to the task difficulty. Comprehensive experiments were conducted on a wide range of LLM benchmarks to verify the presence of their finding and to study the properties that can facilitate its occurrence.

\section{Method}

In this section, we introduce our method and how it is implemented in different algorithms. It is similar to the method described in \citep{li2024agents}. Our method consists of a two-phase process: sampling and voting. During the sampling phase, we try to generate \( N \) samples from \( N \) different agents. After generating these samples, the most popular answer is chosen for each intersection. In the end, as a result, we have an array of actions, each of which was chosen by majority voting.

\subsection{Problem Statement}

During our experiments, there were 12 (for the Jinan dataset) or 16 (for the Hangzhou dataset) intersections with 4 different phases as mentioned in 3. Below is a picture showing the layout of an intersection.

\begin{figure}[h]
    \centering
    \includegraphics[width=0.5\textwidth]{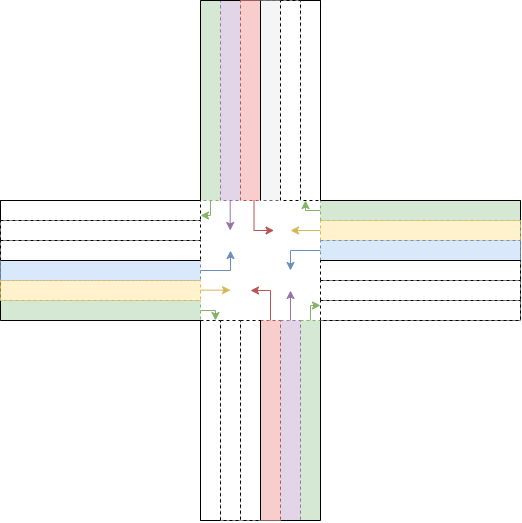} % Укажите имя файла без расширения
    \caption{Intersection scheme}
    \label{fig:example}
\end{figure}

Lanes with the same colour belong to the same phase. It's worth noting that a right turn on the right lane (green colour on the image) is always allowed that's why there're 4 phases.
Our aim in this research is to minimize average queue length (AQL), average travel time (ATT), and average waiting time (AWT). Connectivity between intersections wasn't taken into account because it's hard for LLMs to generalize information from several intersections.

Consider an environment \(\mathcal{E}\) with \( K \) states \(s \in S\) corresponding to each of \( K \) intersections and actions \(a \in A\). In the context of a traffic signal control problem, the environment can be understood as the simulated or real-world traffic system that the control algorithm interacts with. A state in this context can be described as a comprehensive representation of the current traffic conditions at each intersection within the network, such as the number of vehicles in each lane, etc. The state should capture all relevant information needed for the traffic control algorithm to make informed decisions.

Let us have \( K \) states corresponding to each of \( K \) intersections \(\{ s_1, \ldots, s_K \} \) $\in$ \( S \). Then at time step \( t \) we will have \(\{ s_{1t}, \ldots, s_{Kt} \} \) $\in$ \( S_t \). In our implementation, the state at each step contains the current phase and the number of vehicles on each lane. There may be 4 different phases at each intersection: 
1. Left-turn from east and west,
2. Left-turn from north and south,
3. Go-through traffic from east and west, or 
4. Go-through traffic from north and south.

At every time step, the agent chooses the optimal action. An action refers to a specific signal phase chosen for traffic signal control. Each action corresponds to activating a particular signal phase, such as allowing go-through traffic from specific directions or enabling left-turn movements. 

Let us have \(N\) agents \(\{A_1, A_2, \ldots, A_N\}\), each of which can propose an action \(a_i\) at each time step \(t\). Our goal is to choose the optimal action \(a^*\) in each state \(s_t\) using a voting method based on the proposed actions from all agents.

Let \( \pi_i(s_{kt}) \) be the policy of the \(i\)-th agent, which determines the action \(a_i\) in state \(s_{kt}\). The policy is a decision-making strategy employed by the model for traffic signal control. At each time step \(t\), each agent \(A_i\) proposes an action \(a_i = \pi_i(s_{kt})\). After receiving all \(N\) proposals, we choose action \(a^*\) as follows:
\[
a^* = \text{argmax}_a \sum_{i=1}^N \mathbb{I}(a_i = a)
\]
where \(\mathbb{I}\) is an indicator function that equals 1 if \(a_i = a\), and 0 otherwise.

\subsection{Process Description}

\begin{algorithm}
\caption{Sampling Phase}
\begin{algorithmic}[1]
\STATE Initialization of agents: Each of $N$ agents $A_i$ is initialized with its policy $\pi_i$.
\FOR{each state \( s_{kt} \) $k$ $\in$ \(\{ 1, \ldots, K \} \)}
    \FOR{each agent $A_i$}
        \STATE Generate a new action $a_i = \pi_i(s_{kt})$
    \ENDFOR
\ENDFOR
\end{algorithmic}
\end{algorithm}

\begin{algorithm}
\caption{Voting Phase}
\begin{algorithmic}[1]
\STATE Initialize actions collector $a$
\FOR{each agent $A_i$}
    \STATE Collect an action $a_i$ proposed by $A_i$ to actions collector $a$
\ENDFOR
\STATE Choosing an action as illustrated in 3.1: Determine the most popular action $a^*$ from a by majority voting.
\STATE Executing the action: Execute the chosen action $a^*$ in the environment \(\mathcal{E}\).
\end{algorithmic}
\end{algorithm}

\subsection{Implementation Differences for agents based on LLM and RL Algorithms}

There are small differences in implementation for algorithms based on Large Language Models (LLM) and Reinforcement Learning (RL).

\subsubsection{LLM-Based Algorithms}

In LLM algorithms, sampling and voting can be conducted based on textual input, where each agent generates a textual response, and the most frequent response is chosen for further actions. In detail, every agent gets a prompt where task and scenario description are contained. It's worth mentioning that the prompt contains for large language models some facts based on common sense like "The traffic congestion is primarily dictated by the early queued vehicles, with the MOST significant impact" because LLMs basically are not supposed to solve these kind of tasks. The prompt is taken the same as in \citep{lai2024llmlight}, the full version of it may be found in appendix A1. The output of the LLM-based agent is primarily textual data in the form of a reasoning trajectory and control action descriptions. For LLM agents multiagent approach can be used besause predictions of every agent aren't deterministic, in our work the temperature hyperparameter is set to 0.1. The temperature parameter modifies the probabilities of the next word in the sequence by scaling the logits (the raw predictions from the model) before applying the softmax function. This output is generated by the agent to provide insights into the decision-making process and the selected actions for traffic signal control at intersections.

\subsubsection{RL-Based Algorithms}

In the realm of Reinforcement Learning (RL), particularly within the multi-agent paradigm, the orchestration of network updates necessitates a meticulously orchestrated preparation of states for each individual agent. This preparatory phase is indispensable, serving as the bedrock upon which subsequent algorithmic operations rest. This was done for each agent in the current paper.

Also in MPlight, unlike the LLM-based approach, training is conducted, learning rate is 0.001. The input state and target value for each agent are prepared during network updates.

Multi-agent approach which is based on the independence of agents can be used because the recommendations of every agent may vary. The reason of varying recommendations is that during the initialization of MPLight agents, the weights are initialized randomly. Also when selecting an action, agents may choose a random action with some probability. When preparing states and target values for each agent, data is randomly selected from memory (storage of previous interactions with the environment).
\subsection{Algorithm pipeline}

We propose to create a multi-agent system by integrating several (1, 5 or 10) large language models (LLMs) or reinforcement learning (RL) models. This approach aims to leverage the strengths of different models, enhance accuracy, and improve overall system performance by distributing tasks among specialized agents.

To create this system we initialized several identical models which were taken from 3.3. These models generate different actions for each state. From each model, we receive a recommended signal. Using the voting method described in Section 3.1, we determine the final action. Each model's recommendation is considered a vote, and the action with the most votes is selected as the final action. This approach helps to aggregate the diverse recommendations from multiple models, ensuring a more robust decision-making process.

The algorithm below illustrates the whole pipeline of the multi-agent version of the method. It includes initialization, action generation, voting, final action selection and metrics calculation. The multi-agent system is designed to handle various traffic conditions by dynamically adjusting the traffic signals based on real-time data. By doing so, the system can adapt to varying traffic patterns throughout the day, ensuring smooth traffic flow even during peak hours. Moreover, the use of multiple agents for each intersection allows for scalability, enabling the system to manage traffic in large urban areas with numerous intersections effectively.

In the algorithm below there's following variables introduced:
\begin{itemize}
    \item \( Q_i \) is the queue length at intersection \( i \).
    \item \( W_i \) is the waiting time at intersection \( i \).
    \item \( T_i \) is the travel time for vehicle i \( i \).
    \item \( n \) is the  number of vehicles.
\end{itemize}

\FloatBarrier
\begin{algorithm}
\caption{Pipeline of the multi-agent version of algorithm}
\begin{algorithmic}[1]
\STATE Initialize the total run count $T$ and minimum action time \( \tau \) from the configuration .
\STATE Reset the environment and obtain the initial for all intersections state $S_0$.

\FOR{each time step $t$ in range of $T$/\( \tau \)}

    \STATE Initialize empty lists for action $A$.
    
    \FOR{each $s_{kt}$, $k$ $\in$ \(\{ 1, \ldots, K \} \)}
        \STATE Get the information about $s_{kt}$.
    \ENDFOR

    \STATE Initialize an empty list for prompts $P$.
    \FOR{each $s_{kt}$, $k$ $\in$ \(\{ 1, \ldots, K \} \)}
        \STATE Generate a prompt $p_k$ from the $s_{kt}$ and append the prompt to $P$.
    \ENDFOR

    \STATE Initialize an empty list for responses $R$
    \FOR{each $s_{kt}$, $k$ $\in$ \(\{ 1, \ldots, K \} \)}
        \STATE Generate response $r_k$ from the LLM agents using $p_k$ by majority voting.
        \STATE Decode the response and append to $R$.
    \ENDFOR
    \STATE Initialize an empty action list $A$
    \FOR{each $r_k$ $k$ $\in$ \(\{ 1, \ldots, K \} \)}
        \STATE Get the signal from response
        \STATE Append the action code corresponding to the signal to $A$.
    \ENDFOR

    \STATE Step the environment with $A$ and set state $S_{t+1}$.
    \STATE Compute
    \[
    \overline{Q} = \frac{1}{K} \sum_{i=1}^{K} Q_i
    \]
    \STATE Compute
    \[
    \overline{W} = \frac{1}{K} \sum_{i=1}^{K} W_i
    \]
\ENDFOR

\STATE Compute\[
T_{\text{total}} = \sum_{i=1}^{n} T_i
\]

\RETURN $T_{\text{total}}, \overline{W}, \overline{Q}$
\end{algorithmic}
\end{algorithm}

\FloatBarrier
\section{Experimental setup}
\subsection{Datasets}
In this research, we have performed a comparison using 2 datasets: Jinan and Hangzhou. For the Jinan dataset there's 12 intersections (4x3), for the Hangzhou dataset there's 16 intersections (4x4). The Jinan dataset was collected from the Dongfeng subdistrict, Jinan, China, the Hangzhou dataset was collected in the Gudang subdistrict, Hangzhou, China.
\subsection{Metrics}
There're 3 metrics that were used to compare methods with a different number of agents: average travel time (ATT) which measures the mean duration that vehicles take to travel from their origins to their respective destinations, average queue length (AQL) which refers to the mean number of vehicles queuing in the road network, average waiting time (AWT) that quantifies the mean time that vehicles spend queuing at intersections within the road network. 
\subsection{Compared methods}

There're 5 methods compared: 2 transportation, 2 based on LLM, and 1 based on reinforcement learning. Below is a short explanation of these methods.

The Fixed Time Algorithm for traffic signal control is one of the simplest and most widely used methods for managing traffic flow. The main idea is that traffic lights switch between phases according to a predetermined schedule, regardless of the current traffic conditions. This algorithm uses fixed schedules to switch traffic light phases. It does not consider current traffic conditions. All traffic lights switch according to the fixed schedule, regardless of the number of vehicles at the intersection. This method is easy to implement and does not require constant monitoring and updating of traffic data, reducing operational costs.

The Max Pressure algorithm is an advanced traffic signal control method designed to optimize traffic flow at intersections by dynamically adjusting signal phases. The core idea is to use real-time information about vehicle queues to make decisions that minimize overall delays and congestion. The algorithm dynamically adjusts the signal phases at each intersection based on real-time traffic conditions, specifically the vehicle queues at each approach. The final decision depends on the calculation of pressure a difference between the incoming and outgoing queues. Unlike some other algorithms, this algorithm does not require prior knowledge of average traffic demands. It operates solely on current traffic data, making it adaptable to real-time conditions.

The MPLight algorithm is an advanced traffic signal control method that combines the principles of the Max Pressure (MP) algorithm with machine learning techniques, specifically reinforcement learning, to improve traffic management. By setting the reward of RL agents to be the same as
max pressure control objective, each local agent is maximizing its own cumulative reward, which further maximizes the
network throughput under certain constraints. Implementing MPLight may require integration with existing traffic management systems and sensors, which could involve additional costs and technical challenges. 

The main idea of the LLMLight algorithm  is to use a Light Language Model (LLM) called LLMLight for optimized traffic signal control. It leverages LLMs to make effective control actions based on reasoning trajectories and aligns with long-term traffic flow goals. The algorithm aims to enhance traffic efficiency by prioritizing lanes with congestion and providing detailed rationales for decision-making in a transparent and explainable manner. This algorithm requires lots of computational capabilities but for the enhanced version of LLM fine-tuned specifically for the TSC problem, LightGPT, results on average are better than for any other method.  
\section{Experimental results}

Below on the table 1 results of performances are introduced. For algorithms based on RL and LLMs to compare their performance with different numbers of agents this number was 1, 5 or 10. The simulation time is 1 hour. For fixed and Max pressure algorithms number of agents is constant and equal to 1 because the larger number of agents doesn't make sense. The best results for each metric and dataset are highlighted in bold.

\begin{table}[h]
    \centering
    \begin{tabular}{|c|ccc|ccc|}
        \hline %  One horizontal line
         Method &  & Jinan &  & & Hangzhou & \\
         & ATT & AQL & AWT  & ATT & AQL & AWT \\
         \hline %  One horizontal line
         Fixed (1 agent) & 481.793 & 491.033 & 70.987 & 616.017 & 301.325 & 73.987\\
         \hline %  One horizontal line
         Max pressure (1 agent) & 282.579 & 170.708 & \textbf{44.532} & 325.329 & 68.992 & 49.601\\
         \hline %  One horizontal line
         MPLight (1 agent) & 307.82 & 215.93 & 97.88 & 345.60 & 84.7 & 81.97\\
         MPLight (5 agent)  & 304.987 & 209.117 & 99.929 & 349.054 & 87.617 & 98.149\\
         MPLight (10 agent)  & 298.79 & 201.20  & 91.795 & 353.183 & 91.342 & 125.905\\
         \hline %  One horizontal line
         Llama 13b (1 agent) & 403.67 & 371.55 & 135.979 & 458.95 & 175.08 & 199.41 \\
         Llama 13b (5 agent)  & 401.12 & 371.91 & 133.524 & 449.04 & 161.375 & 159.192\\
         Llama 13b (10 agent)  & 404.92 & 370.575 & 139.23 & 441.44 & 160.901 & 199.45\\
         \hline %  One horizontal line
         LightGPT (1 agent) & 277.259 & 164.275 & 46.657 & 312.413 & 58.433 & 47.004\\
         LightGPT (5 agent) & 277.29 & 163.5 & 45.692 & 312.059 & 57.65 & 47.882 \\
         LightGPT (10 agent) & \textbf{274.854} & \textbf{159.775} & 45.847 & \textbf{311.576} & \textbf{57.53} & \textbf{46.079}\\
         \hline %  One horizontal line
    \end{tabular}
    \caption{Performances of models with different number of agents}
    \label{tab:my_label}
\end{table}
\section{Experimental analysis}

As we can see from the table above for some datasets and models multiagent approach shows good results. 

For the MPLight algorithm there's a good advance in results for the Jinan dataset. The reason of it is that in the Hangzhou dataset number of intersections is more than in the Jinan dataset. For the Jinan dataset, the number of intersections is one-third less than that of the Hangzhou dataset. In large intersection datasets, the number of interactions and situations increases, which complicates the learning task. With a higher number of agents at a single intersection, the complexity grows exponentially, which can lead to a deterioration in results. 

For the LLMLight algorithm based on Llama 13b proposed method doesn't show any enhancement. The reason is that this model was not trained for traffic signal control problem and it is too weak for this kind of task. Because of it during conducting experiments the dispersion of Llama 13b results was bigger than on any other methods. Unlike Llama13b LightGPT model shows some improvement in most of all metrics with an increasing number of agents, it showed stable values of metrics. 

In the context of LLMs, it is important to note that their effectiveness in specific management tasks heavily depends on the quality of their pre-training. Models specifically trained on data related to transportation systems and traffic signal control can show significantly better results. Therefore, investing in the training of specialized models can be a strategically justified decision to improve the performance of traffic management systems. This is evident when comparing the results of Llama 13b and LightGPT, which is essentially a fine-tuned version of Llama 13b.

All in all, our results may be summarized using 3 points:

1. The multi-agent approach for RL algorithms shows good results for datasets with a small number of intersections.

2. For "weak" LLMs multiagent approach doesn't show any enhancement because of their instability in results.

3. For fine-tuned LLMs multiagent approach allows to reach small advance in all metrics.

All in all, for cities that have lots of computational capabilities multi-agent approach with a fine-tuned LightGPT model can show the best results among all possible options. If there's not enough computational resources Max pressure algorithm shows good results on all datasets.
\section{Conclusion}

In this paper, we've successfully created multi-agent system and conducted experiments with it for different RL and LLM-based algorithms. Our experiments showed that for some cases especially for fine-tuned LLM models, there's some enhancement when several agents are used. In future works we are going to provide more experiments with different RL-based and LLM-based algorithms. Also, we are going to try this approach with different types of agents and compare it with the results from this paper.

In conclusion, the choice of approach and model for traffic signal control should be based on an analysis of the number of intersections and available computational resources to ensure optimal system performance. Considering the specifics of each city, adapting the approach may include tuning algorithms, improving data quality, and utilizing advanced distributed learning methods. The development of more sophisticated models and algorithms may also involve integrating real-time data, which will allow for more accurate responses to changing traffic conditions and enhance the overall efficiency of the management system.
\newpage
\bibliographystyle{plain}  % Указание стиля библиографии
\bibliography{manuscript}% Указание файла с библиографическими данными (без расширения .bib)
\newpage
\appendix
\section{1}
A traffic light regulates a four-section intersection with northern, southern, eastern, and western sections, each containing two lanes: one for through traffic and one for left-turns. Each lane is further divided into three segments. Segment 1 is the closest to the intersection. Segment 2 is in the middle. Segment 3 is the farthest. In a lane, there may be early queued vehicles and approaching vehicles traveling in different segments. Early queued vehicles have arrived at the intersection and await passage permission. Approaching vehicles will arrive at the intersection in the future. The traffic light has 4 signal phases. Each signal relieves vehicles' flow in the group of two specific lanes. The state of the intersection is listed below. It describes: 

- The group of lanes relieving vehicles' flow under each signal phase.

- The number of early queued vehicles of the allowed lanes of each signal.

- The number of approaching vehicles in different segments of the allowed lanes of each signal.

$<<$There is information about current state$>>$ 

Please answer:
Which is the most effective traffic signal that will most significantly improve the traffic 
condition during the next phase?
Requirements:

- Let's think step by step.

- You can only choose one of the signals listed above.

- You must follow the following steps to provide your analysis: 

Step 1: Provide your analysis 
for identifying the optimal traffic signal.

Step 2: Answer your chosen signal.

- Your choice can only be given after finishing the analysis.

- Your choice must be identified by the tag: $<$signal$>$YOUR\_CHOICE$<$/signal$>$.

%croos section karlie
%GOOD: http://www.eumetrain.org/satmanu/CMs/TrCyAt/print.htm 
%https://physics.stackexchange.com/questions/275799/why-is-the-eye-of-a-cyclone-a-forced-vortex
%http://www.chanthaburi.buu.ac.th/~wirote/met/tropical/textbook_2nd_edition/navmenu.php_tab_9_page_7.1.0.htm
%http://www.atmos.umd.edu/~dalin/andrew/part2.html
%https://nptel.ac.in/courses/119102007/2
%http://www.911omissionreport.com/steering_hurricanes.html
%https://www.youtube.com/watch?v=_brY_9ME8iE brooks
\end{document}